\def\year{2025}\relax
\documentclass[letterpaper]{article} % DO NOT CHANGE THIS
\usepackage{aaai25}  % DO NOT CHANGE THIS
\usepackage{times}  % DO NOT CHANGE THIS
\usepackage{helvet}  % DO NOT CHANGE THIS
\usepackage{courier}  % DO NOT CHANGE THIS
\usepackage[hyphens]{url}  % DO NOT CHANGE THIS
\usepackage{graphicx} % DO NOT CHANGE THIS
\usepackage{natbib}  % DO NOT CHANGE THIS AND DO NOT ADD ANY OPTIONS TO IT
\usepackage{caption} % DO NOT CHANGE THIS AND DO NOT ADD ANY OPTIONS TO IT
\DeclareCaptionStyle{ruled}{labelfont=normalfont,labelsep=colon,strut=off} % DO NOT CHANGE  THIS
\usepackage{xcolor}

\usepackage{subfigure} % 
\usepackage{booktabs} 
\usepackage{amsfonts} 
\usepackage{bm}
\usepackage{algorithm}
\usepackage{algorithmic}
\usepackage{soul}
\usepackage{newfloat}
\usepackage{listings}
\usepackage[section]{placeins}
\usepackage{appendix}
\usepackage{amsmath}
\usepackage{subfigure} 
\usepackage{float}
\usepackage{subcaption}
\usepackage{lipsum}
\graphicspath{{./images/}}
\floatstyle{ruled}
\newfloat{listing}{tb}{lst}{}
\floatname{listing}{Listing}
\title{Adaptive Gradient Regularization: A Faster and Generalizable Optimization Technique for Deep Neural Networks}
\author{
    Huixiu Jiang\textsuperscript{1,2,3} , 
    Ling Yang\textsuperscript{5},
    Yu Bao\textsuperscript{6},
    Rutong Si\textsuperscript{4}\footnotemark[4]
    Sikun Yang\textsuperscript{1,2}\footnotemark[4]
}
\affiliations{
    \textsuperscript{\rm 1}School of Computing and Information Technology, Great Bay University, 523000 Dongguan, China
    \\
    \textsuperscript{\rm 2}Great Bay Institute for Advanced Study, Great Bay University \\
    \textsuperscript{\rm 3}Department of Computer Science, Illinois Institute of Technology Chicago, USA\\
    \textsuperscript{\rm 4}The Abdus Salam International Centre for Theoretical Physics, Strada Costiera 11, Trieste 34151, Italy\\
    \textsuperscript{\rm 5}School of Computer Science, Peking University, China \\
    \textsuperscript{\rm 6}Bioinformatics Center, Institute for Chemical Research, Kyoto University, Japan\\
    huixiujiang23@gmail.com, yangling0818@163.com, bitdanger51@gmail.com, rsi@ictp.it, sikunyang@gmail.com
}
\begin{document}
\maketitle
\linespread{0.85}
\begin{abstract}
Stochastic optimization plays a crucial role in the advancement of deep learning technologies. Over the decades, significant effort has been dedicated to improving the training efficiency and robustness of deep neural networks, via various strategies including gradient normalization (GN) and gradient centralization (GC). Nevertheless, to the best of our knowledge, no one has considered to capture the optimal gradient descent trajectory, by adaptively controlling gradient descent direction. 
To address this concern, this paper is the first attempt to study a new optimization technique for deep neural networks, using the sum normalization of a gradient vector as coefficients, to dynamically regularize gradients and thus to effectively control optimization direction. The proposed technique is hence named as the adaptive gradient regularization (AGR). It can be viewed as an adaptive gradient clipping method. The theoretical analysis reveals that the AGR can effectively smooth the loss landscape, and hence can significantly improve the training efficiency and model generalization performance. We note that AGR can greatly improve the training efficiency of vanilla optimizers' including Adan and AdamW, by adding only three lines of code. The final experiments conducted on image generation, image classification, and language representation,  demonstrate that the AGR method can not only improve the training efficiency but also enhance the model generalization performance. 
\end{abstract}
\section{1.~ ~ Introduction}
\noindent Over the decades, deep learning has achieved great success in promoting technological and scientific advancements in broad fields including computer vision~\cite{he2016deep,lecun2015deep}, natural language processing~\cite{devlin2018bert,vaswani2017attention}, scientific computing~\cite{wang2023scientific}, and etc. On the one hand, the wide development of large-scale datasets~\cite{feifeili} and abundant computing resources including GPUs and TPUs, greatly benefit the advancement of deep learning technologies. On the other hand, elegant neural network architectures~\cite{LIU201711} and stochastic gradient descent algorithms~\cite{Bottou2012}, ensure deep neural networks to be able to capture large-scale datasets, with satisfying training efficiency and excelllent learning performance.\\
\indent Recent studies on the optimization techniques for neural networks, aim to accelerate the neural network training efficiency and also to improve the model generalization performance. To the end, there has been a growing number of stochastic gradient-based algorithms, among which stochastic gradient descent (SGD)~\cite{Ruder, Bottou2012} is the earliest and most representative one for its simplicity and efficiency in training many sophisticated network architectures. Afterward, many variants of SGD have been dedicated to improving training efficiency and learning performance, e.g. SGD with momentum (SGDM)~\cite{QIAN1999145}, Adam~\cite{Kingma2014}, AdamW~\cite{Loshchilov2017}, Adagrad~\cite{Duchi}, RMSProp~\cite{tieleman2012lecture} and ACProp~\cite{zhuang2021}. In particular, the introduction of first-order and second-order momentum combined with adaptive learning strategies, enables us to effectively train deep neural networks (DNNs).\\
\indent  
Notably, some recent studies~\cite{chengn, Qiao} find that the statistics of gradients can be utilized to regularize gradients, and hence to stabilize the training of neural networks. 
The most prevalent method is normalization, e.g., batch normalization (BN)~\cite{Ioffe}, group normalization (GN)~\cite{wu}, layer normalization (LN)~\cite{ba2016layer}, and instance normalization (IN)~\cite{Dmitry}. BN utilizes the statistics of samples in a mini-batch to approximate the statistics of the entire training set, to circumvent the internal covariate shift issue. As a result, the training speed can be significantly improved, without sacrificing the learning performance. %, much higher learning rates, and less care about initialization. %\\
% \indent 
BN has been incorporated into the design of many network architectures, and its ideology greatly benefits the subsequent studies~\cite{salimans,huang2017, Qiao}. 
%Many network architectures have incorporated BN as the indispensable layer, and the ideology it invents provides insights for subsequent research. 
For instance, inspired by BN, weight normalization (WN)~\cite{salimans} decouples the length of weight vectors from their direction, and hence can effectively speed up the full batch normalization in training deep neural networks. 
Weight standardization (WS)~\cite{Qiao} regularizes the weights to have zero mean and unit variance, to smooth the optimization landscape by reducing the Lipschitz constants of the loss function and gradients. Inspired by WN and WS, gradient centralization (GC)~\cite{Yong} is proposed to operate on the gradients of weight, by simply centralizing the gradient vectors to have zero mean, which can regularize both the weight space and ambient feature space, to boost the generalization performance of DNNs.\\
\indent Despite making some progress in accelerating neural network training efficiency, these optimization techniques still fail to adaptively control gradient descent direction, which may lead to finding the \emph{sub-optimal} gradient descent trajectories. To mitigate this gap, we thoroughly study a novel optimization technique to adjust the gradient descent direction \emph{dynamically}. To the end, we adaptively regularize the gradients using the sum normalization of a gradient vector as regularization coefficients, and hence can better approximate the \emph{optimal} gradient descent trajectory. More specifically, we provide both the theoretical analysis and extensive experimental results to validate our hypothesis and the efficacy of the developed optimization technique% and theoretical analysis provides a precise guarantee
. The main contributions of our work are as follows:\\%Despite being greatly successful in accelerating the training process and improving the generalization performance through gradient normalization and centralization, there's still less attention to optimization direction control in the gradient descent. We conjecture large differences between gradients' magnitude can lead to deviating from the optimal path. To mitigate the limitation mentioned above, we propose a novel optimization technique to adjust the gradient descent direction dynamically. Our experimental results certify our hypothesis and theoretical analysis provides a precise guarantee. The main contributions of our work are as follows:\\
\begin{itemize}
\vspace{-4mm}
\item A novel gradient regularization strategy, is proposed to dynamically control the gradient descent direction.
\item A new optimization technique is developed by adaptively regularizing gradients on the \emph{element-scale}, to improve the training efficiency and model generalization performance for deep neural networks. We hence call it the Adaptive Gradient Regularization (AGR). In particular, theoretical analysis are provided to guarantee that the AGR can effectively accelerate the training efficiency and improve the model generalization performance for deep neural networks, by smoothing the loss landscape.
\item Extensive experiments conducted on various tasks including image generation, image classification, and language representation, demonstrate the superior performance of the proposed AGR, in terms of improving the training efficiency for deep neural networks. 
\end{itemize}

\section{2.~ ~ Related Work}
\textbf{Weight:} Weight Normalization (WN) 
~\cite{salimans,gitman2017} is a simple reparameterization of the weight vectors in DNNs that accelerates the convergence of SGD optimization, which is an alternative to Batch Normalization (BN). Weight Standardization (WS)~\cite{Qiao} is targeted at the micro-batch training where the small batch sizes are not enough for training DNNs with Batch Normalization (BN), which normalizes weight vectors instead of features like Batch Normalization and can smooth the loss landscape by reducing the Lipschitz constants of the loss and the gradients. \\
\textbf{Gradient:} Gradient is calculated for loss back-propagation in deep network training to update weight. Various gradient improvement methods have been proposed for a more stable training process and better generalization performance. Momentum is defined as the first-order momentum and the second-order momentum of gradients to be introduced as optimizer development method~\cite{QIAN1999145,7966082}. In SGDM, momentum helps SGD accelerate in the relevant direction and dampen oscillations by considering the previous updates~\cite{Ruder}. SGD with Nesterov Acceleration~\cite{Nesterov,tran2022nesterov} takes into account the momentum term in the calculation of the gradient, by considering the future position of the parameter. Gradient clipping~\cite{Zhang2019WhyGC,chen2020understanding,menon2019can} and gradient normalization~\cite{chengn,wu2021gradient} were proposed for training convergence acceleration. Gradient clipping is to solve the gradient exploding problem by scaling the gradient to keep it small when the gradient becomes too large, improving training stability and accelerating convergence. gradient normalization is similar to weight normalization decoupling the length of gradients from their direction. Gradient Centralization (GC)~\cite{yong2020gradient} centralizes gradients to have zero means, which can be viewed as a projected gradient descent method with a constrained loss function and regularizes the weight space and output feature space boosting the generalized performance of DNNs.\\ 
\indent Distinguished from weights and activations, the distribution of gradients is approximately lognormal\cite{chmiel2020neural,guo2021partition}. The distribution has a few gradients with huge magnitude and many gradients with small magnitude. The logarithmic scale of such distribution is close to a symmetric normal (Gaussian) distribution around the mean as zero, which is consistent with the assumption that gradient in the neural network is sampled from Gaussian distribution~\cite{Wiedemann2020DitheredBA}. During the training process, the gradient distribution should be maintained as unchanged as possible. In other words, the difference between gradients needs to be as small as possible. That is why gradient clipping was proposed to accelerate DNN training. However, the current gradient modification methods could not adjust each gradient according to its magnitude dynamically. It is challenging to set an appropriate threshold in gradient clipping.  \\
\textbf{Adaptive learning rate:} Determining a good learning rate can prevent the system from diverging in terms of the loss function and slow learning~\cite{chen2018}. Adagrad~\cite{Duchi} scales the learning rate adaptively in each iteration for all dimensions based on the sum of the outer product of the gradients, considering all previous updates. However, the learning rate will be almost unchanged if the gradient is large at the beginning and small subsequently. RMSprop~\cite{tieleman2012lecture} introduces decay factors in the outer product of the gradients instead of simply summing them, solving the problem of Adagrad. Adam~\cite{Kingma2014} and AdamW~\cite{Loshchilov2017} also adapt the ideology to scale the learning rate adaptively. Adan~\cite{xie2022adan} is an adaptive Nesterov algorithm reformulating the vanilla Nesterov momentum to develop a new Nesterov momentum estimation (NME), which improves the model training speed across multiple networks.\\
\indent In this paper, we propose a gradient regularization technique in the element scale based on their magnitude in an adaptive way.
\section{3~ ~ Adaptive Gradient Regularization}
%\subsection{3.1 Methodology}
\noindent Normalization and standardization have been applied on features and weight vectors, respectively, to speed up neural network training, such as BN~\cite{Ioffe}, WN~\cite{salimans,gitman2017}, and WS~\cite{Qiao}, etc. These methods can reduce the Lipschitz constant of the loss function and smooth the optimization landscape. Apart from applying to samples and weight vectors, GC~\cite{Yong} computes the mean of gradients, and then centralizes the gradients to have zero mean, instead of performing standardization on gradients. GC can effectively improve the Lipschitzness of the loss function, and thus can improve the generalization performance, with a strong restraint by projecting the gradient on a hyperplane. The statistical operation on gradients by gradient normalization or centralization in multi-dimensions, does not consider the gradient magnitude on the element scale, and thus distributes the fixed coefficient to all the gradients. It may lead to capturing sub-optimal gradient descent trajectories since we cannot dynamically control the optimization direction. To address this limitation, we introduce the adaptive coefficient for each gradient by the sum normalization of the gradient vector in a neural network layer. Then, each gradient subtracts the product of itself and the corresponding coefficient, to adaptively approximate the optimal gradent descent direction.
\subsection{3.1 Notations}
\noindent %There are some notations to be defined. 
We consider the gradients of the weight matrix in convolutional layers and fully connected layers. For convolutional layers, we denote the weight matrix (kernel) by $\textbf{W}_{\mathcal{C}} \in \mathbb{R}^{\mathcal{C}_{in}\times\mathcal{C}_{out}\times\mathcal{K}\times\mathcal{K}}$, in which $\mathcal{C}_{in}$ is the number of input channels, $\mathcal{C}_{out}$ is the number of output channels, and $\mathcal{K}\times\mathcal{K}$ is the size of convolutional kernel. For fully connected layers, we denote the weight matrix by $\textbf{W}_{\mathcal{F}} \in \mathbb{R}^{\mathcal{C}_{in}\times\mathcal{C}_{out}}$, in which $\mathcal{C}_{in}$ is the number of input channels, and $\mathcal{C}_{out}$ is the number of output channels. For simplicity, we denote the weight matrix $\textbf{W} \in \mathbb{R}^{\mathcal{M}\times\mathcal{N}}$ in an unified format. In addition, $\textbf{w}_{i,j}$ is the $(i,j)$-th entry 
%element in \textit{i-th} row and \textit{j-th} column 
of $\textbf{W}$. Denote \textbf{W} the weight matrix, and $\mathcal{L}$ the objective function. Hence,  $\nabla_{\textbf{w}} \mathcal{L}$ and $\nabla_{\textbf{w}_{i,j}} \mathcal{L}$ denote the gradient of $\mathcal{L}$ w.r.t. the weight matrix \textbf{W} and the weight $\textbf{w}_{i,j}$, respectively. We use $\left| \nabla_{\textbf{w}} \mathcal{L} \right|$ to denote the absolute value of $ \nabla_{\textbf{w}} \mathcal{L}$. Let $\sum \nabla_{\textbf{w}_{i,j}} \mathcal{L}$ to be the sum of  $\nabla_{\textbf{w}_{i,j}} \mathcal{L}$ over all the dimensions.
\subsection{3.2 AGR Formulation}
\noindent For the fully connected layer and convolutional layer, suppose the gradient of weight $\textbf{w}_{i,j}$ is $\nabla_{\textbf{w}_{i,j}} \mathcal{L}(i=1,2,...M;j=1,2,...N)$. We denote the AGR operator by $\Psi$ and give the definition as below: 
\begin{equation}
\Psi(\nabla_{\textbf{w}_{i,j}} \mathcal{L})=\nabla_{\textbf{w}_{i,j}} \mathcal{L}-\alpha_{i,j} \nabla_{\textbf{w}_{i,j}} \mathcal{L},
\end{equation}
\begin{equation}
\alpha_{i,j}=\frac{\left| \nabla_{\textbf{w}_{i,j}} \mathcal{L} \right|}{\sum_{i,j}\left| \nabla_{\textbf{w}_{i,j}} \mathcal{L}\right|}.
\end{equation}
The adaptive coefficient $\alpha_{i,j}$ is the ratio of the  absolute value of the gradient to the sum of the  absolute values of the gradients over all the dimensions.% \textit{i} and \textit{j}. 
This method considers the gradient magnitude in the element scale, and thus can assign an adaptive coefficient to each gradient, to ensure adaptive gradient regularization. The weight update in each iteration using AGR is% ()
\begin{equation}
\textbf{w}^{t+1} = \textbf{w}^{t}-\eta\Psi(\nabla_{\textbf{w}_{i,j}} \mathcal{L}),
\end{equation}
where $\eta$ is the learning rate.

\subsection{3.3 Applying AGR to AdamW/Adan Optimizers}
\noindent It is convenient to embed AGR into AdamW~\cite{Loshchilov2017} and Adan~\cite{xie2022adan} with just three lines of code. More specifically, we update the weights by replacing $\nabla_{\textbf{w}_{i,j}} \mathcal{L}$ with $\Psi(\nabla_{\textbf{w}_{i,j}} \mathcal{L})$ to add AGR into AdamW and Adan as shown in Algorithm 1 and Algorithm 2, respectively. We also found no additional computational cost and training time in multiple neural networks. Notably, we only apply the AGR to the gradients and the first-order momentum instead of the second-order momentum because we expect to reduce the learning rate and regularize the gradient consistently when the gradient is large.
\section{4. AGR Properties}
\noindent In this section, we shall provide a theoretical analysis to explain why the studied adaptive gradient regularization can accelerate the training efficiency and improve the generalization performance of deep neural networks in different tasks.

\begin{algorithm}[htbp]
\caption{Adan with AGR}

\textbf{Input}: initialization $\theta_{0}$, step size $\eta$, momentum ($\beta_{1}$,$\beta_{2}$,$\beta_{3}$)$\in [0,1]^{3}$, stable parameter $\epsilon>0$, weight decay $\lambda_{k}>0$,restart condition. $\mathbf{m_{0}} = \mathbf{g_{0}}$, $\mathbf{v_{0}}=0$, $\mathbf{v_{1}} = \mathbf{g_{1}}-\mathbf{g_{0}}$, $\mathbf{n_{0}} = \mathbf{g_{0}^{2}}$. \\
\textbf{Output}: some average of ${\left\{\bm{\theta_{k}}\right\}}_{k=1}^{K}$
\begin{algorithmic}[1] %[1] enables line numbers
\WHILE{$k<K$}
\STATE calculate the stochastic gradient $\mathbf{g_{k}}$ at $\bm{\theta_{k}}$;\\
$\overline{\mathbf{g_{k}}} = \Psi(\mathbf{g_{k}})$;\\
$\mathbf{m_{k}} = (1-\beta_{1})\mathbf{m_{k-1}} + \beta_{1}\overline{\mathbf{g_{k}}}$;\\
$\mathbf{v_{k}} = (1-\beta_{2})\mathbf{v_{k-1}} + \beta_{2}(\overline{\mathbf{g_{k}}}-\mathbf{g_{k-1}})$;\\
$\mathbf{n_{k}} = (1-\beta_{3})\mathbf{n_{k-1}} + \beta_{3}[\mathbf{g_{k}}+(1-\beta_{2})(\mathbf{g_{k}}-\mathbf{g_{k-1}})]^{2}$;\\
$\bm{\eta_{k}}=\eta/(\sqrt{\mathbf{n_{k}}}+\epsilon)$;\\
$\bm{\theta_{k+1}}=(1+\lambda_{k}\eta)^{-1}[\bm{\theta_{k}}-\bm{\eta_{k}}\circ(\mathbf{m_{k}}+(1-\beta_{2})\mathbf{v_{k}})]$;\\
\IF {\textit{restart condition holds}}
\STATE estimate stochastic gradient $\mathbf{g_{0}}$ at $\bm{\theta_{k+1}}$;\\
set $k=1$ and update $\bm{\theta_{1}}$ by Line 6;
\ENDIF
\ENDWHILE
\end{algorithmic}
\end{algorithm}

\begin{algorithm}[htbp]
\caption{AdamW with AGR}
\textbf{Input}: given $\alpha=0.001, \beta_1=0.9, \beta_2=0.999, \epsilon=10^{-8}, \lambda \in \mathbb{R}$. \\
\textbf{Parameters}: initialize time step $t \leftarrow 0$, parameter vector $\boldsymbol{\theta}_{t=0} \in \mathbb{R}^n$, first moment vector $\boldsymbol{m}_{t=0} \leftarrow \boldsymbol{0}$, second moment
    vector $\boldsymbol{v}_{t=0} \leftarrow \boldsymbol{0}$, schedule multiplier $\eta_{t=0} \in \mathbb{R}$\\
\textbf{Output}: optimized parameters $\boldsymbol{\theta}_t$
\begin{algorithmic}[1] %[1] enables line numbers
\REPEAT
\STATE$t \leftarrow t+1$;\\
$\nabla f_t\left(\boldsymbol{\theta}_{t-1}\right) \leftarrow$ SelectBatch
$\left(\boldsymbol{\theta}_{t-1}\right)$;\\
$\boldsymbol{g}_t \leftarrow \nabla f_t\left(\boldsymbol{\theta}_{t-1}\right)+\lambda \boldsymbol{\theta}_{t-1}$;\\
$\overline{\boldsymbol{g_{t}}} = \Psi(\boldsymbol{g_{t}})$;\\
$\boldsymbol{m}_t \leftarrow \beta_1 \boldsymbol{m}_{t-1}+\left(1-\beta_1\right) \overline{\boldsymbol{g_{t}}}$;\\ 
$\boldsymbol{v}_t \leftarrow \beta_2 v_{t-1}+\left(1-\beta_2\right) \boldsymbol{g_{t}}^2$;\\
$\hat{\boldsymbol{m}}_t \leftarrow \boldsymbol{m}_t /\left(1-\beta_1^t\right)$; \\
$\hat{\boldsymbol{v}}_t \leftarrow v_t /\left(1-\beta_2^t\right)$ \\
$\eta_t \leftarrow SetScheduleMultiplier (t)$ \\
$\boldsymbol{\theta}_t \leftarrow \boldsymbol{\theta}_{t-1}-\eta_t\left(\alpha \hat{\boldsymbol{m}}_t /\left(\sqrt{\hat{\boldsymbol{v}}_t}+\epsilon\right)+\lambda \boldsymbol{\theta}_{t-1}\right)$
\UNTIL stopping criterion is met
\end{algorithmic}
\end{algorithm}

\begin{figure*}[!ht] 
    \centering
    \subfigure[\label{fig:a}]{
        \includegraphics[width=5.6cm,height=3.7cm]{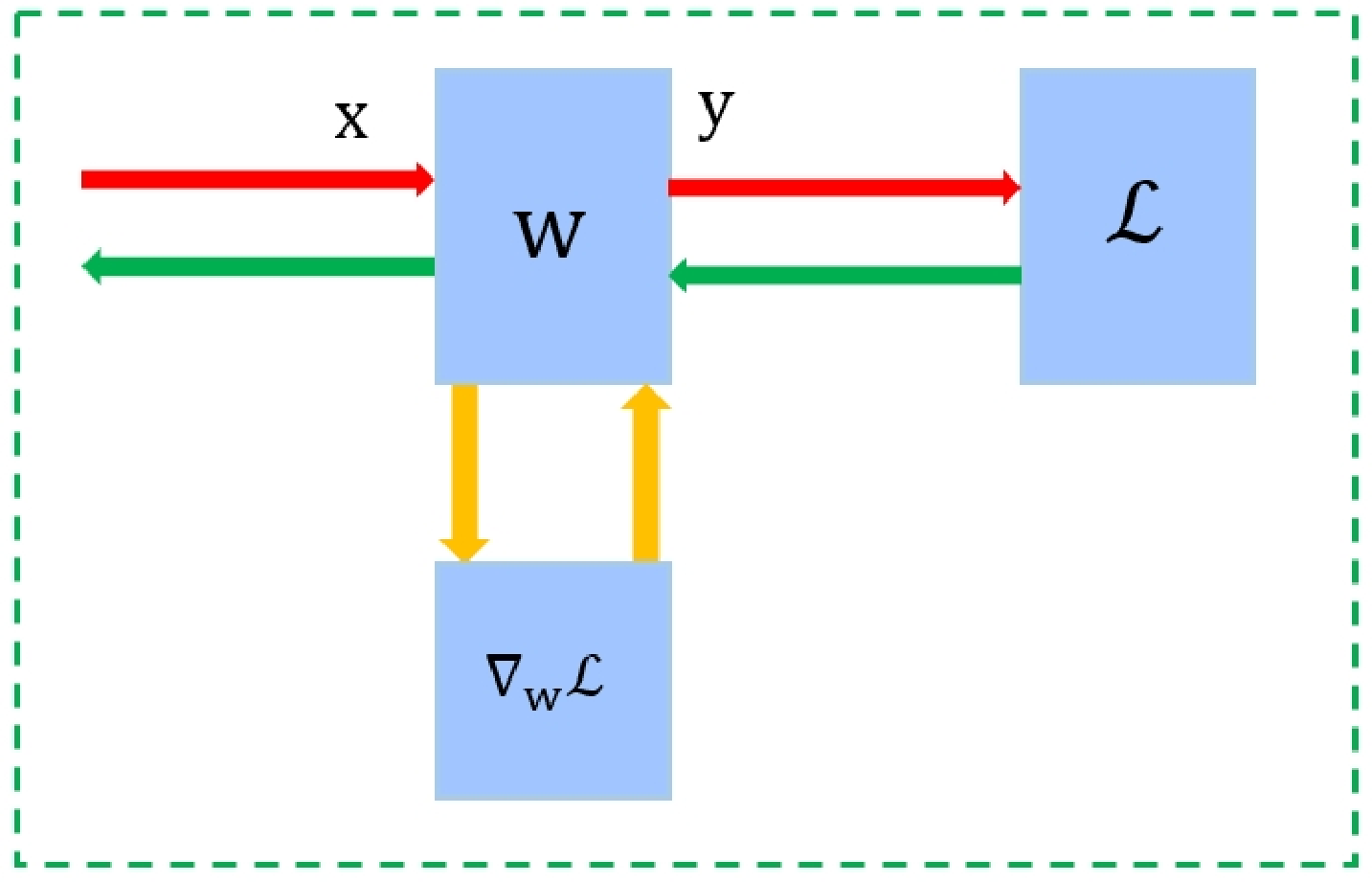}
    }
    \subfigure[\label{fig:b}]{
        \includegraphics[width=5.6cm,height=3.7cm]{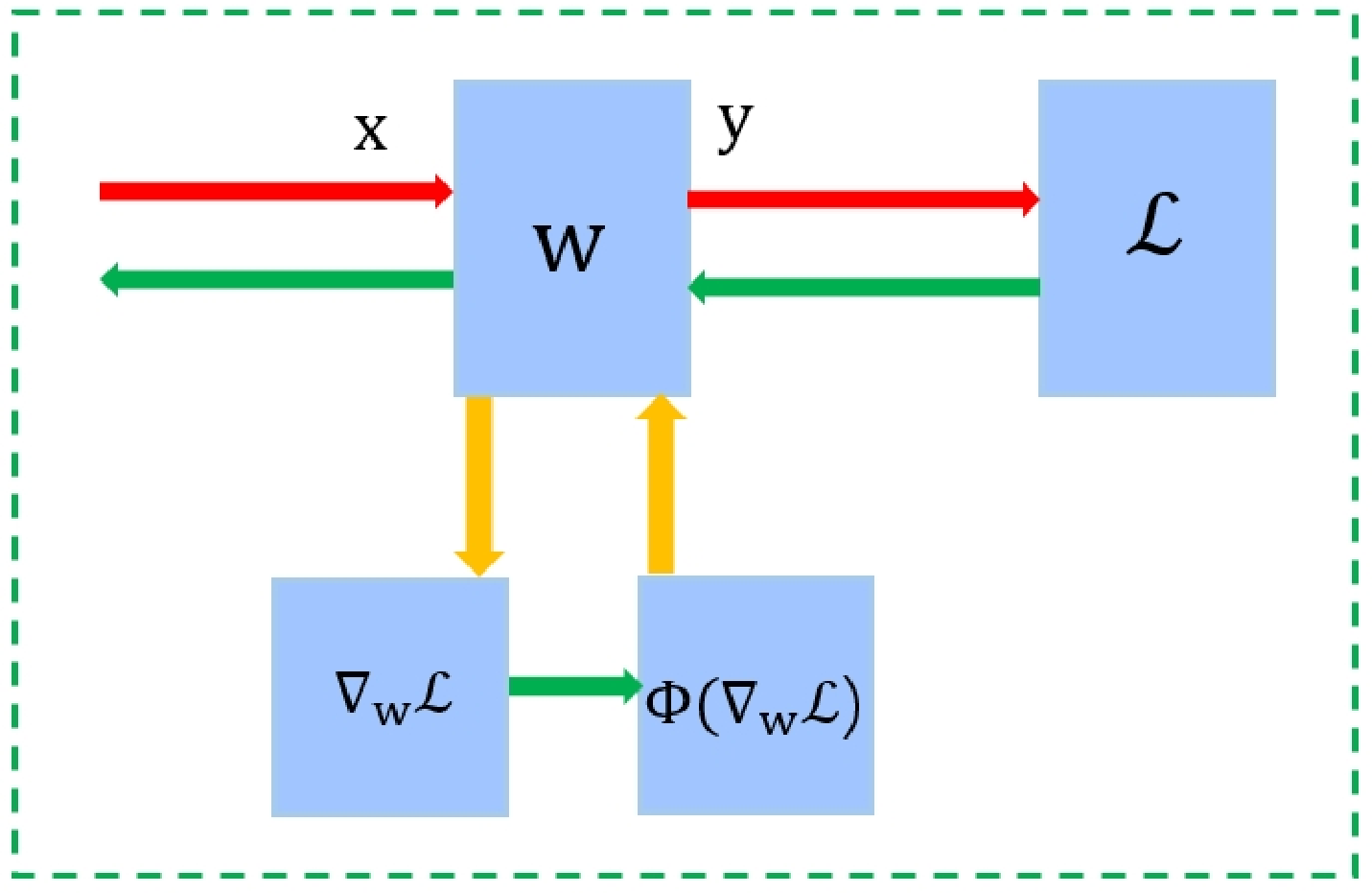}
    }
    \subfigure[\label{fig:c}]{
        \includegraphics[width=5.6cm,height=3.7cm]{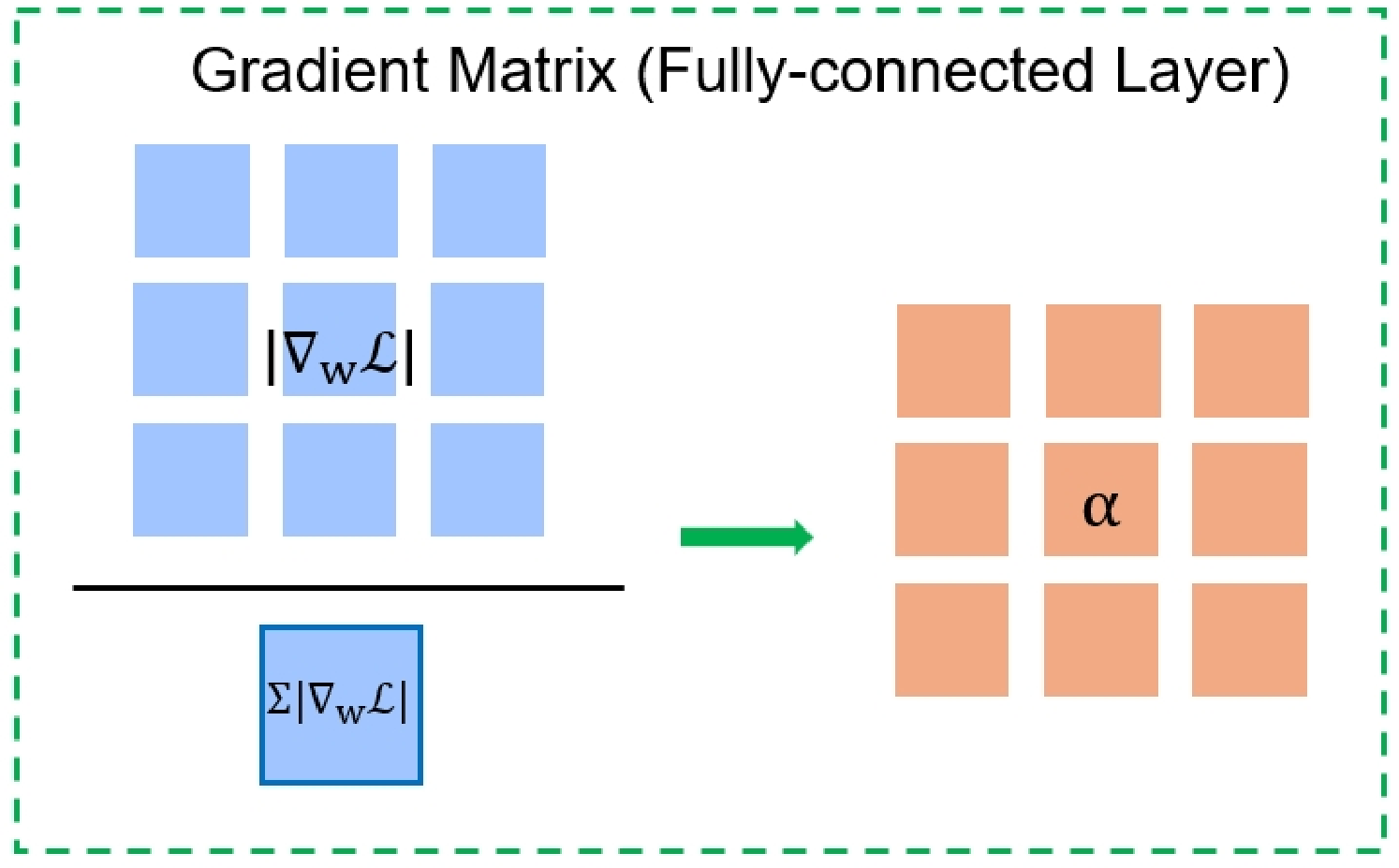}
    }
    \vspace{-3mm}
    \caption{(a),(b) are sketches of how the AGR is embedded into the vanilla optimizer. \textbf{W} is the weight tensor, $\mathcal{L}$ is the loss function, $\nabla_{w}\mathcal{L}$ is the gradient of weight, and $\Psi(\nabla_{w}\mathcal{L})$ is the gradient with AGR method. (c) is the sketch of the AGR 
    calculation,$\left|\nabla_{w}\mathcal{L}\right|$ is the absolute value of the gradient, $\sum\left|\nabla_{w}\mathcal{L}\right|$ is the sum of $\left|\nabla_{w}\mathcal{L}\right|$ w.r.t all dimensions. The black line represents the ratio, we can obtain the corresponding coefficient matrix.}
    \vspace{-3mm}
\end{figure*}

\noindent\textbf{Smoothing the loss landscape:}  
%Loss landscape represents optimization performance during the training process, which indicates training stability and model robustness. Lipschitzness of the loss and its gradient can evaluate the loss landscape. 
The loss landscape of deep neural network provides a valuable perspective for investigating the training stability, robustness and generalization of neural networks. Note that we can evaluate the loss landscape by the Lipschitzness and gradient of the loss function. In particular,  
%Specifically, 
we use the magnitude of $\left|\left|\nabla_{\textbf{w}}\mathcal{L}\right|\right|_{2}$ and $\left|\left|\nabla_{\textbf{w}}^{2}\mathcal{L}\right|\right|_{2}$to capture the Lipschitzness of loss and its gradient, respectively. 
We also note that a better Lipschitzness of the loss with smaller $\left|\left|\nabla_{\textbf{w}}\mathcal{L}\right|\right|_{2}$, often indicates a smoother landscape and more effective training performance.\\
%Less magnitude indicates a smoother loss landscape and better optimization performance.\\
\textbf{Theorem 4.1} \textit{$\nabla_{\textbf{w}}\mathcal{L}$ is the gradient of loss function $\mathcal{L}$ w.r.t. weight $\textbf{w}$, with the function $\Psi(\nabla_{\textbf{w}} \mathcal{L})$ defined in Eq.(1). We have the following results about the loss function and its gradient}:
\begin{equation}  
\left\{  
     \begin{array}{lr}  
     \left|\left|\Psi(\nabla_{\textbf{w}}\mathcal{L})\right|\right|_{2} \leq \left|\left|\nabla_{\textbf{w}}\mathcal{L}\right|\right|_{2}, &  \\  \left|\left|\nabla_{\textbf{w}}\Psi(\nabla_{\textbf{w}}\mathcal{L})\right|\right|_{2} \leq \left|\left|\nabla_{\textbf{w}}^{2}\mathcal{L}\right|\right|_{2}. & \\    
     \end{array}  
\right.  
\end{equation}
\noindent The proof of theorem 4.1 is attached in the \textbf{Appendix}. 
% It shows that the loss $\mathcal{L}$ and its gradient $\nabla_{\textbf{w}}\mathcal{L}$ are restrained with the AGR method. It reveals that our method provides better Lipschitzness by implementing a dynamic regularization on the gradients.
Intuitively, the loss $\mathcal{L}$ and its gradient $\nabla_{\textbf{w}}\mathcal{L}$ using AGR, can be restricted by a tighter bound. Hence, we naturally can improve the Lipschitzness of the loss function and its gradient in training deep neural networks, by dynamically regularizing gradients with AGR. 
\\ 
\textbf{Adjusting learning rate using gradients:} During the update of weight, the learning rate and gradient contribute to the update direction and magnitude together~\cite{Cutkosky_2023}. The first-order momentum and second-order momentum are introduced to accelerate loss descending, and to rescale the learning rate, respectively. In addition, the learning rate can also be affected by the gradient magnitude. In other words, the learning rate for a weight at each iteration, can be adaptively adjusted according to its gradient, using the AGR method. 
% From this perspective, our AGR method also integrates an adaptive learning rate with a specific gradient. 
\\
\textbf{Theorem 4.2} \textit{AGR adaptively adjusts the learning rate $\eta_{i,j}$ using the gradient $\nabla_{\textbf{w}_{i,j}}\mathcal{L}$. 
%their adaption tendency keep consistent.
}\\
The proof of theorem 4.2 is attached in the \textbf{Appendix}. Note that AGR can simutaneously regularize the gradients and effectively adjust the learning rate, via Eq.(2). Intuitively, AGR regularizes the gradient when it is too large, and thus naturally decreases the learning rate, accordingly.
% we can understand that the operation on gradients and the learning rate are not separate. It is a better way to view why the AGR works well. 
We thus expect to achieve faster training speed, and better generalization performance by obtaining the smoother loss landscape using the AGR method.

\section{5.~ ~ Experimental Results}
\noindent 
% We apply our AGR method to vision tasks and natural language processing (NLP) tasks. 
We evaluate the effectiveness of the AGR method in improving the training efficiency and generalization performance for neural networks, using various tasks including image classification, image generation, and language representation. 
%Vision tasks consist of image generation and classification. 
For the image generation, we applied the AGR method to speed up the training of the U-Net architecture~\cite{Ronneberger} of the denoising diffusion probabilistic model (DDPM)~\cite{ho2020denoising}. For the image classification, we mainly evaluated the AGR method under the conventional supervised setting, e.g., in speeding up the training of CNN architectures: ResNets~\cite{he2016identity}, VGG~\cite{iglovikov2018}, ConvNext~\cite{liu2022convnet} and Transformer architectures: ViT~\cite{yin2022vit,yuan2021tokens} and Swin~\cite{liu2021swin,liang2021swinir}. For the language representation task, we applied the AGR method in speeding up the training of ALBERT~\cite{devlin2018bert,lan2019albert}. %\\
% \indent 
More specifically, we evaluate the effectiveness of the AGR in improving the training efficiency and generalization performance of neural networks, by applying the AGR to the state-of-the-art optimizers, and comparing the improvements brought by the AGR. 
% We focus on applying AGR to the SOTA optimizer in the architectures mentioned above to evaluate their performance. 
For example, In the image classification task, the default/SOTA optimizer is AdamW~\cite{Loshchilov2017} in CNN-type architectures and ViTs. For the NLP task, the default/SOTA optimizer is LAMB, and we set the optimizer as AdamW. In addition, we also applied the AGR method to Adan optimizer~\cite{xie2022adan}, which achieves better performance than AdamW in training U-Net architecture for the image generation. 
% Restricted by the computational resource, we do not consider to implement the AGR modification in all the optimizers like SGD and Adam for these tasks. 
We considered to apply AGR to AdamW, and to compare it with AdamW instead of Adam and SGD, because SGD and Adam achieve much worse performance than AdamW, for image classification and generation.
\begin{table*}[!ht]
\centering
\caption{\centering{IS and its standard deviation(std) among different optimizers in DDPM} on CIFAR10}
\vspace{-4mm}
\fontsize{9}{9}\selectfont{\begin{tabular}{ccccccc}
\toprule
IS$\uparrow$ & 10 & 400& 800& 1200&  1600& 2000 \\
\midrule
ACProp & 2.95 (0.03) & 7.48 (0.07) & 8.30 (0.07) & 8.44 (0.09) & 9.18 (0.16) & 9.83 (0.17) \\
RMSprop & 3.92 (0.04) & 6.94 (0.07) & 8.62 (0.14)  & 8.61 (0.05) & 9.25 (0.12) & 9.30 (0.06) \\
Adam & 3.76 (0.04) & 7.73 (0.13) & 8.44 (0.11) & 8.96 (0.16) & 9.01 (0.09) & 9.16 (0.13) \\
AdamW & 3.99 (0.04) & 7.84 (0.08) & 8.91 (0.12) & 9.02 (0.04) & 9.11 (0.09) & 9.18 (0.15) \\
Adan & 4.31 (0.06) & 8.14 (0.13) & 8.85 (0.10) & 9.18 (0.10) & 9.19 (0.07) & 9.22 (0.11) \\
\textbf{Adan(AGR)} & \textbf{4.38(0.05)} & \textbf{8.32 (0.10)} & \textbf{8.86 (0.12)} & \textbf{9.18 (0.08)} & \textbf{9.26 (0.13)} & \textbf{9.34 (0.12)} \\
\bottomrule
\end{tabular}
}
\vspace{-2mm}
\end{table*}
\begin{table}[!ht]
\captionsetup{singlelinecheck=off}
\caption{FID score on CIFAR10 dataset among optimizers}
\vspace{-3mm}
\fontsize{9}{9}\selectfont{\begin{tabular}{p{1.5cm}p{0.7cm}p{0.6cm}p{0.6cm}p{0.6cm}p{0.6cm}p{0.8cm}}
\toprule
FID$\downarrow$ & 10 & 400& 800& 1200& 1600& 2000 \\
\midrule
ACProp & 304.03 & 40.06 & 28.42 & 12.32 & 10.12 & 9.57 \\
RMSprop & 313.19 & 54.67 & 23.56 & 10.35 & 8.40 & 8.10 \\
Adam & 229.30 & 55.12 & 21.91 & 10.45 & 10.27 & 9.27 \\
AdamW & 210.46 & 23.90 & 17.43 & 10.05 & 9.11 & 9.01 \\
Adan & 169.32 & 14.60 & 13.68 & 8.64 & 8.07 & 7.98 \\
\textbf{Adan(AGR)} & \textbf{161.65} & \textbf{13.48} & \textbf{12.75} & \textbf{8.08} & \textbf{7.85} & \textbf{7.44} \\
\bottomrule
\end{tabular}
}
\vspace{-5mm}
\end{table}
\linespread{0.83}
\subsection{5.1 Experimental Setup}
\noindent To thoroughly evaluate the AGR method, we considered to conduct the experiments in training various neural architectures, using multiple representative datasets.

% across image generation, image classification, and NLP tasks across multiple popular neural networks and open-source datasets. 
More specifically, the experimental setup are detailed as follows:
\begin{itemize}
    \item We conducted image generation task using the CIFAR10 dataset~\cite{krizhevsky2010}, which consists of 10 classes, and each class has 5000 images for training and 1000 for testing. The image resolution is 32 $\times$ 32. In this task, we used the famous diffusion model paradigm: Denoising Diffusion Probabilistic Model (DDPM) with U-Net architecture as a solver for training and applied the AGR to the Adan optimizer. Moreover, we also included the other popular optimizers in the comparison.
    \item We conducted image classification using multiple classical neural networks (VGG, ResNet, ViTs, ConvNext, etc.), on the CIFAR100 dataset~\cite{lin2013network}, and Tiny-ImageNet~\cite{le2015tiny}. For the experiments, we trained the tiny version of ViTs and ConvNext. The CIFAR100 dataset consists of 100 classes, each of which has 500 images for training and 100 for testing. The image resolution is 32 $\times$ 32. The Tiny-ImageNet consists of 200 classes, each of which has 500 images for training, 50 for validating, and 50 for testing. The image resolution is 64 $\times$ 64. As the labels of the images are not provided in the test set, we thus use the validation dataset for evaluation.
    \item Finally, we conducted language representation tasks on the WikiText-2 dataset~\cite{radford2019language}. For the experiments, we trained ALBERT~\cite{lan2019albert}, a light version of BERT, for masked language model (MLM) and sentence order prediction (SOP) tasks.
    \item For the better evaluation, each of the experiments is conducted four times, and then we calculated the mean to report the final results.
\end{itemize}

Unless otherwise specified, we applied the AGR method to the fully connected layers and convolutional layers of the neural architectures in the experiments. For Adan and AdamW, we used their default settings while using the LambdaLR learning rate scheduler if comparison experiments are not conducted on the hyperparameters. There is no integrated and additional operation on the optimizers except for the simple three lines of codes representing the AGR method. All the experiments are conducted on NVIDIA A40 with Pytorch framework(2.0 version).

\begin{table*}[ht]
\centering
\caption{\centering{Top-1 ACC.(\%) of ConvNext and Swin on Tiny-ImageNet under the official settings}}
\vspace{-3mm}
\fontsize{9}{9}\selectfont{\begin{tabular}{ccccccccccccc}
\toprule
 &  \multicolumn{3}{c}{ConvNext Tiny}& \multicolumn{3}{c}{ConvNext Tiny}&  \multicolumn{3}{c}{Swin Tiny}& \multicolumn{3}{c}{Swin Small} \\
Epoch & 100 & 200& 300& 100 & 200& 300 & 100 & 200& 300& 100 & 200& 300\\
\midrule
AdamW  &64.78 & 67.47& 68.47& 65.92 & 67.81& 70.02& 64.79 & 69.85& 72.01& 65.43 & 70.49& 72.12 \\
\textbf{AdamW(AGR)} & \textbf{64.98} & \textbf{67.71}& \textbf{69.03}& \textbf{66.12} & \textbf{68.90}& \textbf{70.54}& \textbf{65.17} & \textbf{70.16}& \textbf{72.21}& \textbf{66.02} & \textbf{70.69}& \textbf{72.39}  \\
\bottomrule
\end{tabular}
}
\end{table*}

\begin{table*}[!ht]
\centering
\caption{\centering{Top-1 ACC.(\%) of ViTs on Tiny-ImageNet under the official setting}}
\vspace{-3mm}
\fontsize{9}{9}\selectfont{\begin{tabular}{cccccccccc}
\toprule
 & \multicolumn{3}{c}{TinyViT-5M}& \multicolumn{3}{c}{TinyViT-11M}& \multicolumn{3}{c}{TinyViT-21M} \\
Epoch & 100 & 200& 300& 100 & 200& 300& 100 & 200& 300 \\
\midrule
AdamW & 61.80 & 68.14& 69.50& 63.91 & 70.25& 71.39& 65.06 & 71.38& 72.50  \\
\textbf{AdamW(AGR)} &\textbf{62.96} & \textbf{68.75}& \textbf{69.83}& \textbf{65.71}& \textbf{72.45}& \textbf{73.02}& \textbf{66.13} & \textbf{71.94}& \textbf{72.96}  \\
\bottomrule
\end{tabular}
}
\end{table*}
\vspace{-2mm}
\begin{figure*}[!ht] 
    \centering
    \subfigure{
        \includegraphics[width=8cm,height=6.2cm]{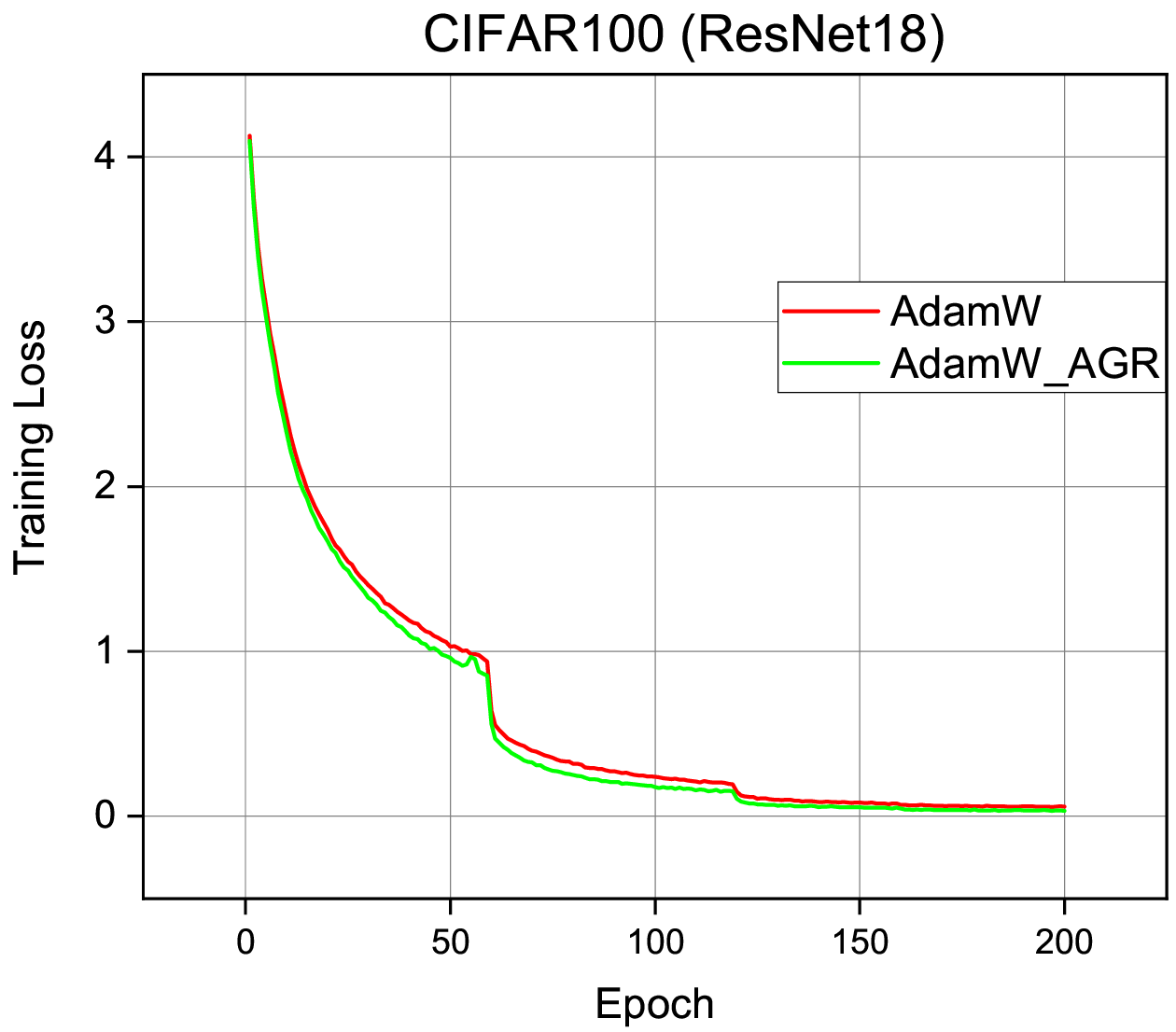}
    }	
        \quad
    \subfigure{
        \includegraphics[width=8cm,height=6.2cm]{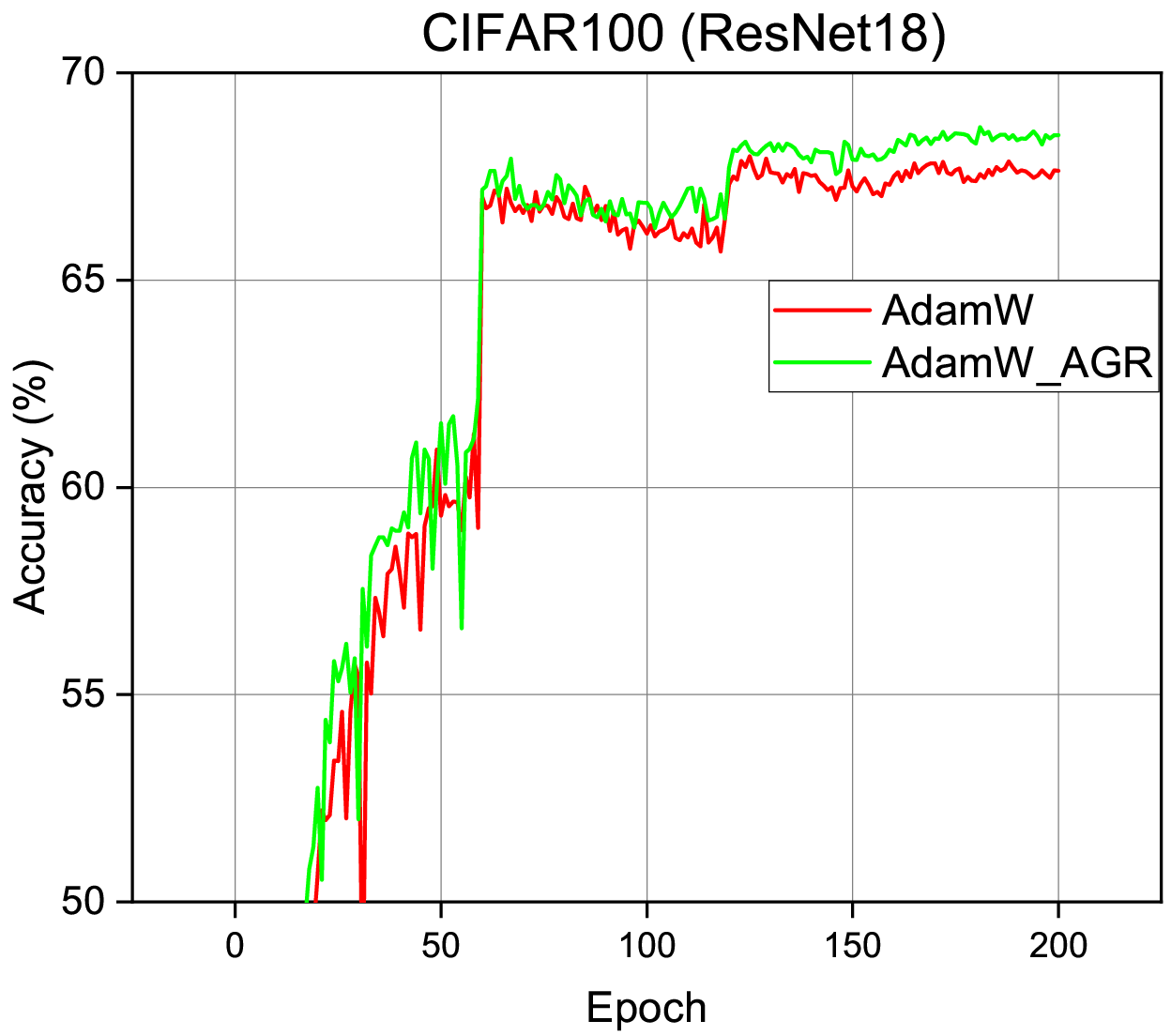}
    }
    \vspace{-1mm}
    \caption{Training loss and test accuracy of ResNet18 structures in Tiny-Imagenet, AGR represents AGR is embedded into AdamW optimizer.}	
    \vspace{-6mm}
\end{figure*}
\vspace{2mm}
\subsection{5.2 Generative Model: DDPM}
\noindent In this section, we present the results of DDPM training on the CIFAR10 dataset using the compared optimizers. We evaluate the performance by FID and IS scores. % respectively., 
We also calculated the standard deviation of the IS score in generating $50,000$ images. For each of the experiments, we trained DDPM with $2,000$ epochs with a batch size being $128$. To save time, we introduce the denoising diffusion implicit model (DDIM)~\cite{song2020denoising} with steps being set to $100$, to accelerate the generation in each evaluation.   
%\indent 
As shown in Table 1 and Table 2, the Adan(AGR) optimizer achieves the best performance among the compared optimizers, which also validates its effectiveness in the image generation task. 
%We note that Adan(AGR) achieves the best performance with a lower FID score and higher IS score. 
It means that the AGR method can improve the learning performance of U-Net architecture, which validates that AGR is an effective optimization technique for image generation. 
We also consider to employ this method in the training of other diffusion models in future work, to validate its effectiveness.  Note that we only consider the comparisons using the AGR method under the same setting. Hence, we omit the hyperparameter tuning, which may lead to achieving better accuracy.
\subsection{5.3 Supervised Classification on TinyImageNet and CIFAR100}
\noindent For the image classficaiton tasks, we applied the AGR to the AdamW optimizer, and evaluated the improvements bround by AGR. For the CIFAR100 dataset, each of the experiments is conducted with $200$ epochs on $2$ GPUs with batch size being $64$ per GPU. For the Tiny-ImageNet dataset, each of the experiments is conducted with $300$ epochs on $2$ GPUs. The batch size is $128$ per GPU.
\begin{table*}[ht]
\centering
\caption{\centering{Top-1 ACC.(\%) of ResNet and VGG11 on Tiny-ImageNet under the official settings}}
\vspace{-3mm}
\fontsize{9}{9}\selectfont{\begin{tabular}{ccccccccccccc}
\toprule
 & \multicolumn{3}{c}{VGG11} & \multicolumn{3}{c}{ResNet-18}& \multicolumn{3}{c}{ResNet-50}& \multicolumn{3}{c}{ResNet-101} \\
Epoch & 50 & 150& 200 & 50 & 150& 200& 50& 150& 200& 50 & 150& 200 \\
\midrule
AdamW & 61.95 & 65.09 & 65.40 & 59.31 & 67.26 &67.64 & 61.26 & 68.19 & 68.80 & 61.47 & 68.61 & 69.33  \\
\textbf{AdamW(AGR)} & \textbf{61.45} & \textbf{65.48} & \textbf{65.73} & \textbf{61.54} & \textbf{67.91} & \textbf{68.50} & \textbf{63.08} & \textbf{68.22}& \textbf{69.38}& \textbf{64.11} & \textbf{68.92}& \textbf{69.73}  \\
\bottomrule
\end{tabular}
}
\end{table*}

\subsection{5.3.1 Results on CIFAR100}
\noindent We evaluated the AGR method in the training of some classical CNN- architectures on the CIFAR100 dataset. 
% We can see the Top-1 accuracy being improved to different degrees. 
We can see that how much the improvments brought by AdamW(AGR) relative to AdamW is, for the Top-1 accurarcy. 
As shown in Figure 2, the training loss goes deeper with the AGR method, and the test accuracy is improved and smoother, which shows that the AGR method can stabilize the training and improve the learning performance for the ResNet18 backbone. We also applied the AGR to AdamW for other popular backbones such as ResNet50 and ResNet101. As shown in  Figure 3, the training loss of VGG11 is lower, and test accuracy is higher, using the AGR method. Hence, it achieves better accuracy in the test datasets in a smoother way, which validates that the AGR can accelerate the training, and improve the generalization performance of AdamW, for the VGG11 backbone. We also present the result of the AGR for the ResNet18 backbone. As shown in Table 3, the AGR method can improve the performance of AdamW with around $1\%$ improvements for the Top-1 accuracy. % of AdamW on them. 

\begin{table}[!ht]
\centering
\caption{SOP ACC.(\%) of Albert on Wiki-Text2 dataset}
\vspace{-3mm}
\fontsize{9}{9}\selectfont{\begin{tabular}{cccccc}
\toprule
Weight decay & 0 & 2e-4& 5e-4& 1e-3 & 1e-2 \\
\midrule
AdamW & 80.02 & 82.14& 83.01& 82.94 & 82.73  \\
\textbf{AdamW(AGR)} &\textbf{80.55} & \textbf{84.19}& \textbf{84.47}& \textbf{83.21}& \textbf{83.06} \\
\bottomrule
\end{tabular}
}
\vspace{-2.5mm}
\end{table}

\begin{table}[!ht]
\centering
\caption{Top-1 ACC.(\%) of TinyViT-11M on Tiny-ImageNet}
\vspace{-3mm}
\fontsize{9}{9}\selectfont{\begin{tabular}{p{1.8cm}p{0.55cm}p{0.55cm}p{0.55cm}p{0.55cm}p{0.55cm}p{0.55cm}}
\toprule
Epochs & 0 & 100& 150& 200& 250& 300 \\
\midrule
\textbf{AdamW(AGR)} &71.39 &\textbf{71.57}&\textbf{72.88}& \textbf{72.92}& \textbf{73.58}& \textbf{73.02} \\
\bottomrule
\end{tabular}
}
\vspace{-2.5mm}
\end{table}

\begin{table}[!ht]
\centering
\caption{FID score of DDPM on CIFAR10}
\vspace{-2.5mm}
\fontsize{9}{9}\selectfont{\begin{tabular}{p{1.8cm}p{0.55cm}p{0.55cm}p{0.55cm}p{0.55cm}p{0.55cm}p{0.55cm}}
\toprule
Epochs & 0 & 400& 800& 1200& 1600& 2000 \\
\midrule
\textbf{Adan(AGR)} &\textbf{7.73} &\textbf{7.64}&\textbf{7.58}& \textbf{7.02}& \textbf{7.81}& \textbf{7.44} \\
\bottomrule
\end{tabular}
}
\vspace{-5.5mm}
\end{table}
\subsection{5.3.2 Results on Tiny-ImageNet}

\noindent We conducted the experiments on the Tiny-ImageNet. To evaluate the efficiency and generalization performance of the AGR, we applied the AGR method to train Transformer architectures and ConvNext using the Tiny-imageNet dataset. For a more comprehensive comparison, we applied the AGR to the multiple variants of TinyViT, Swin, and ConvNext. Specifically, there are three variants of TinyVit with 5M, 11M, and 21M parameters. 
%There are also the tiny versions of ConvNext and Swin, respectively. 
As shown in Figure 3, the AGR method can further decrease the training loss, and improves the test accuracy, through the entire training. The test accuracy of Swin-Tiny is improved with a large margin in the later epochs. 
%, which also improves the generalization capability and accelerates the training. 
As shown in Table 3, the AGR method can improve the performance with \textgreater$0.5\%$ improvement, for the ConvNext and Swin backbones. As shown in Table 4, the test accuracy for the TinyViT variants, is improved by 1.7\% at most. Similar to the multiple ResNet variants, TinyViT with more parameters cannot also excavate more pattern information due to the limited training images which achieves a little higher accuracy than TinyViT with fewer parameters. Hence, the AGR method can be embedded into multiple neural networks in training the Transformer.
\begin{figure*}[!ht] 
    \subfigure{
        \includegraphics[width=5.5cm,height=5.9cm]{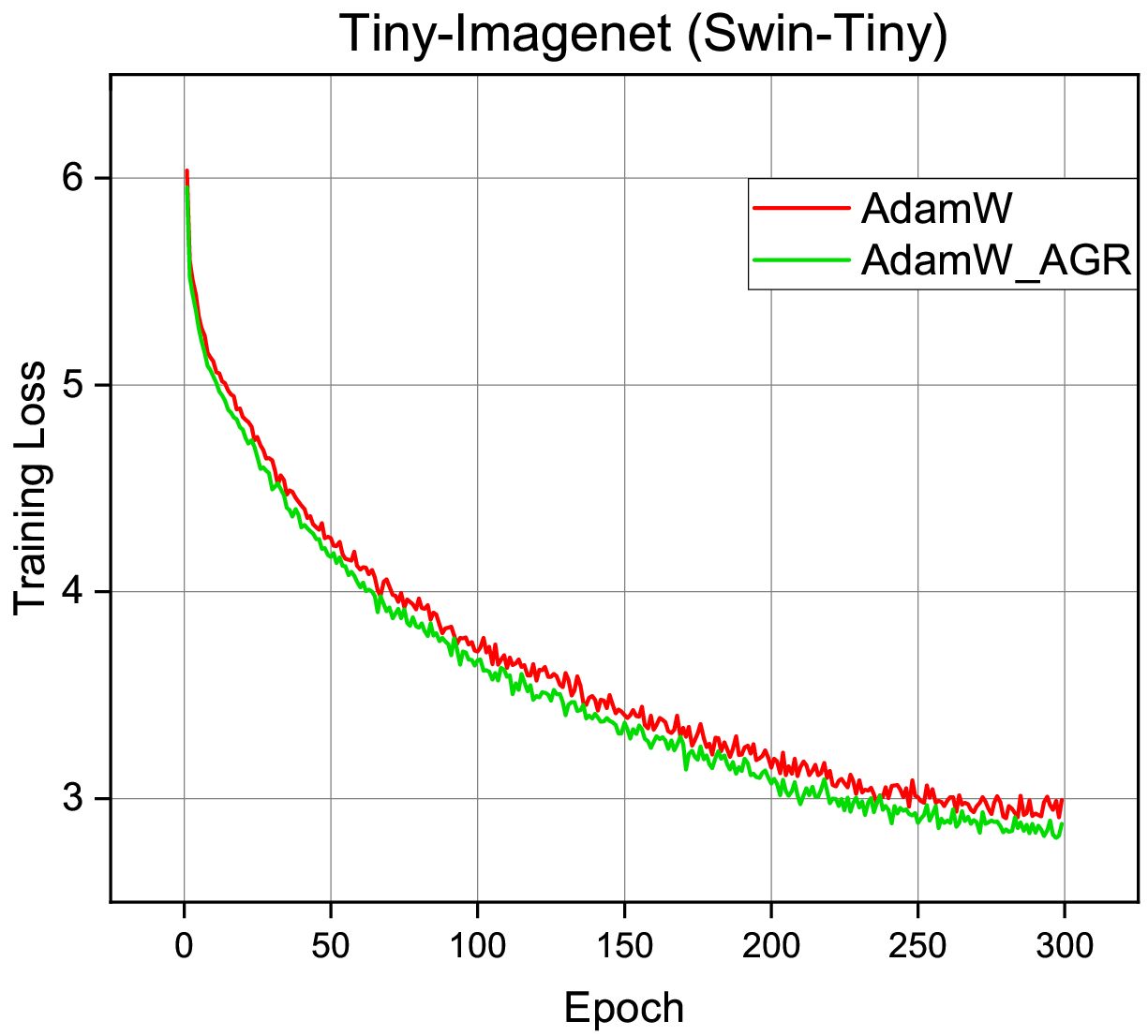}
    }	
        %\quad
    \subfigure{
        \includegraphics[width=5.5cm,height=5.9cm]{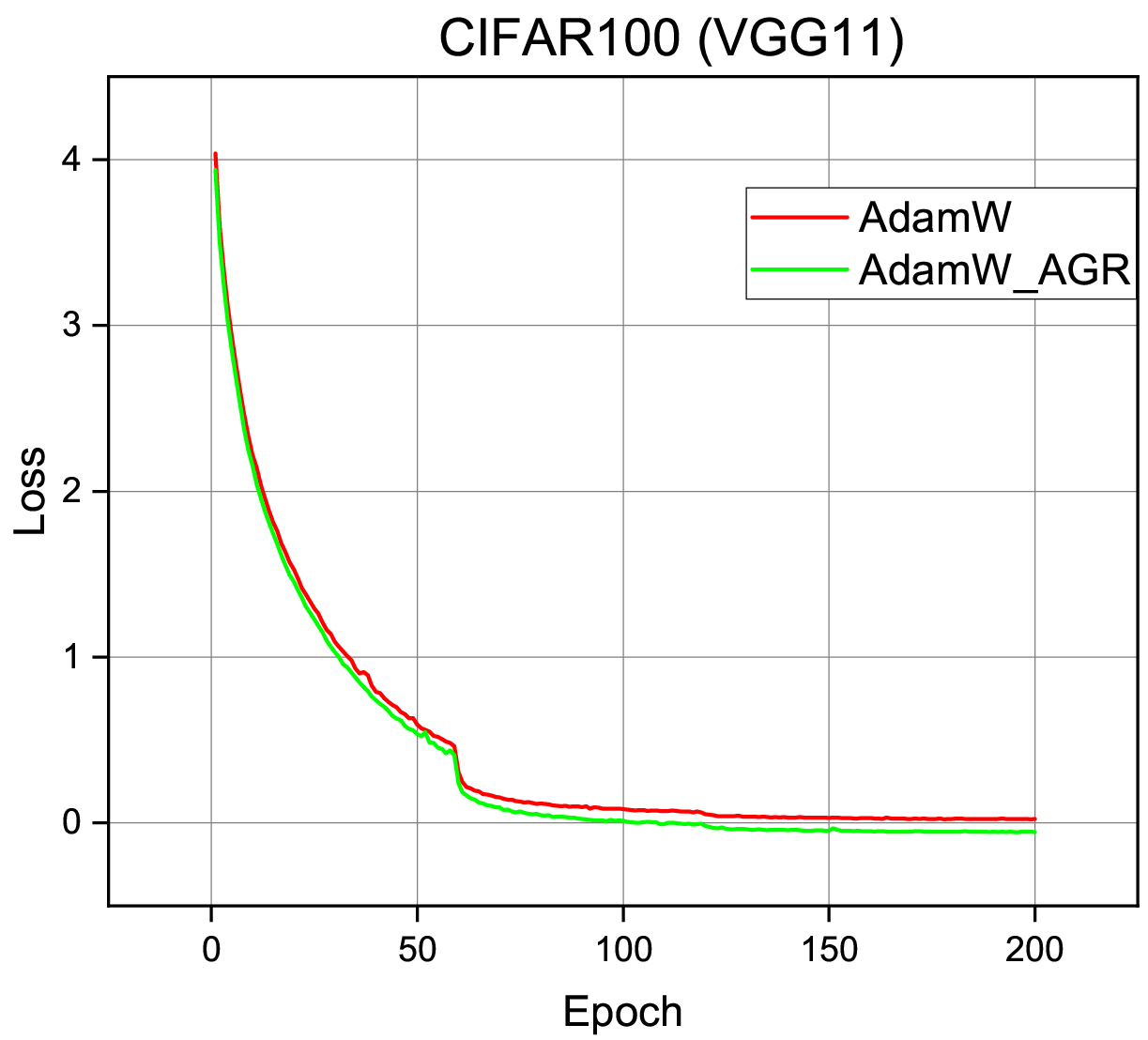}
    }	
        %\quad
    \subfigure{
        \includegraphics[width=5.5cm,height=5.9cm]{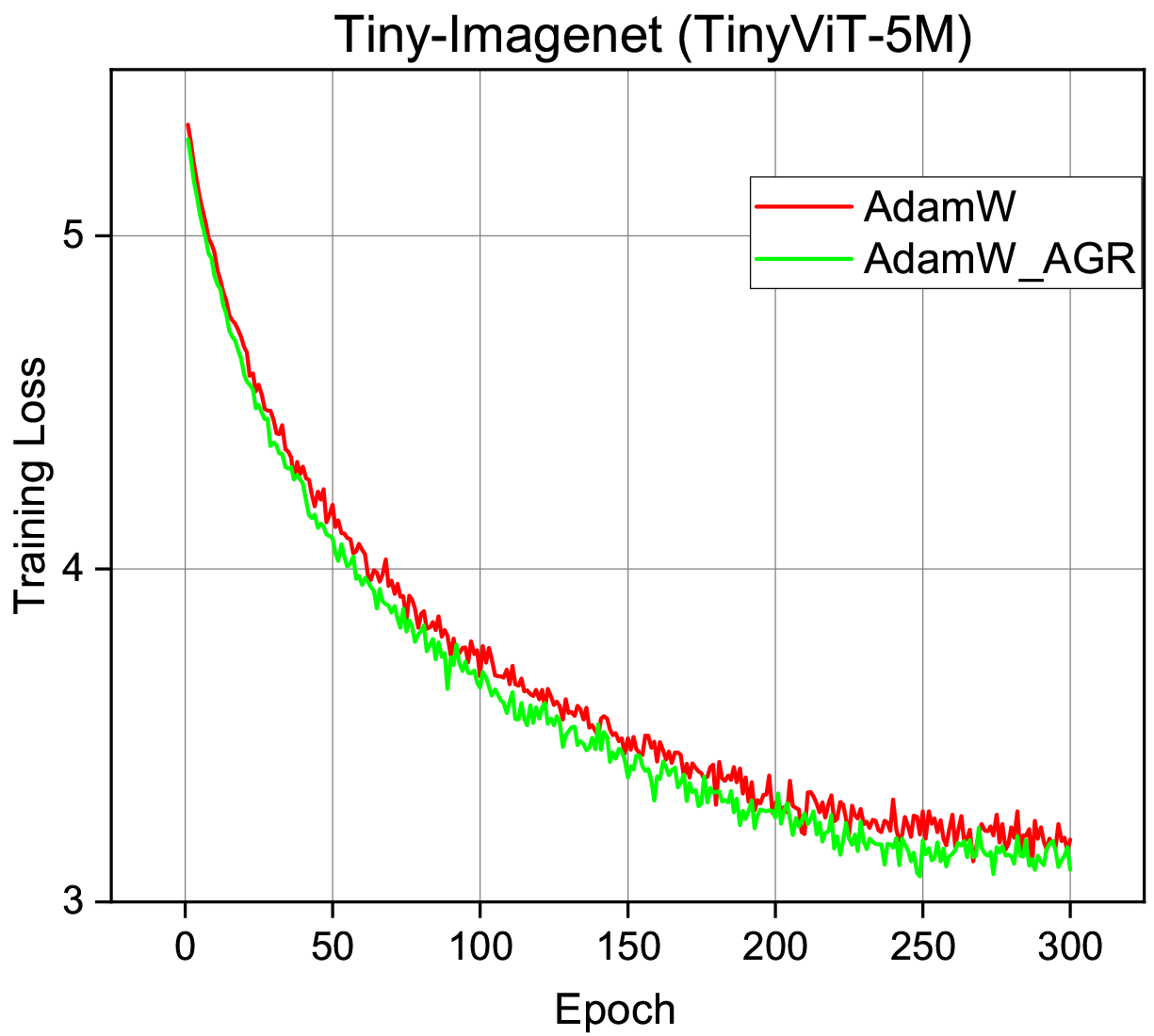}
    }	
        %\quad
    \subfigure{
        \includegraphics[width=5.5cm,height=5.9cm]{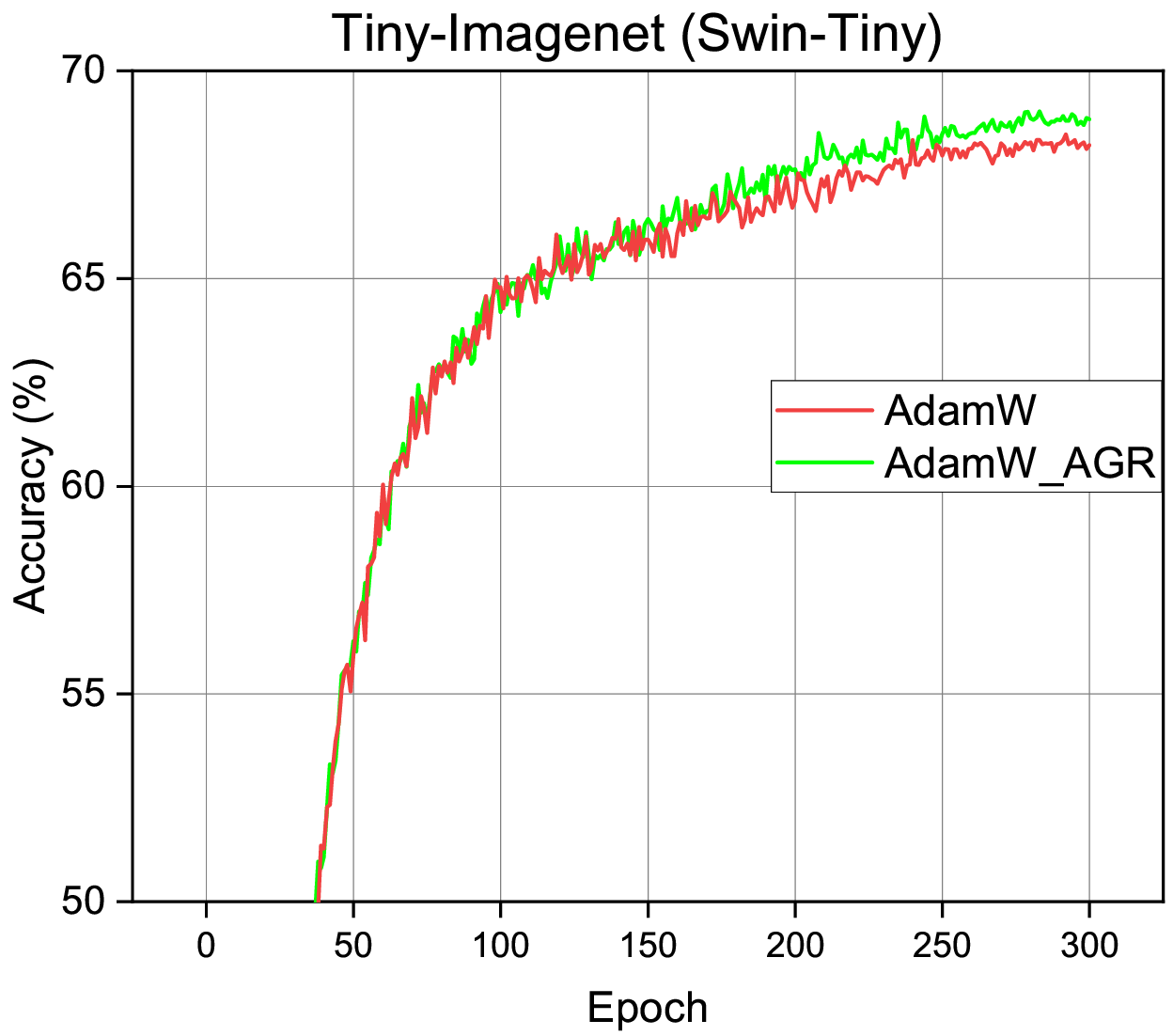}
    }	
    \quad
    \subfigure{
        \includegraphics[width=5.5cm,height=5.9cm]{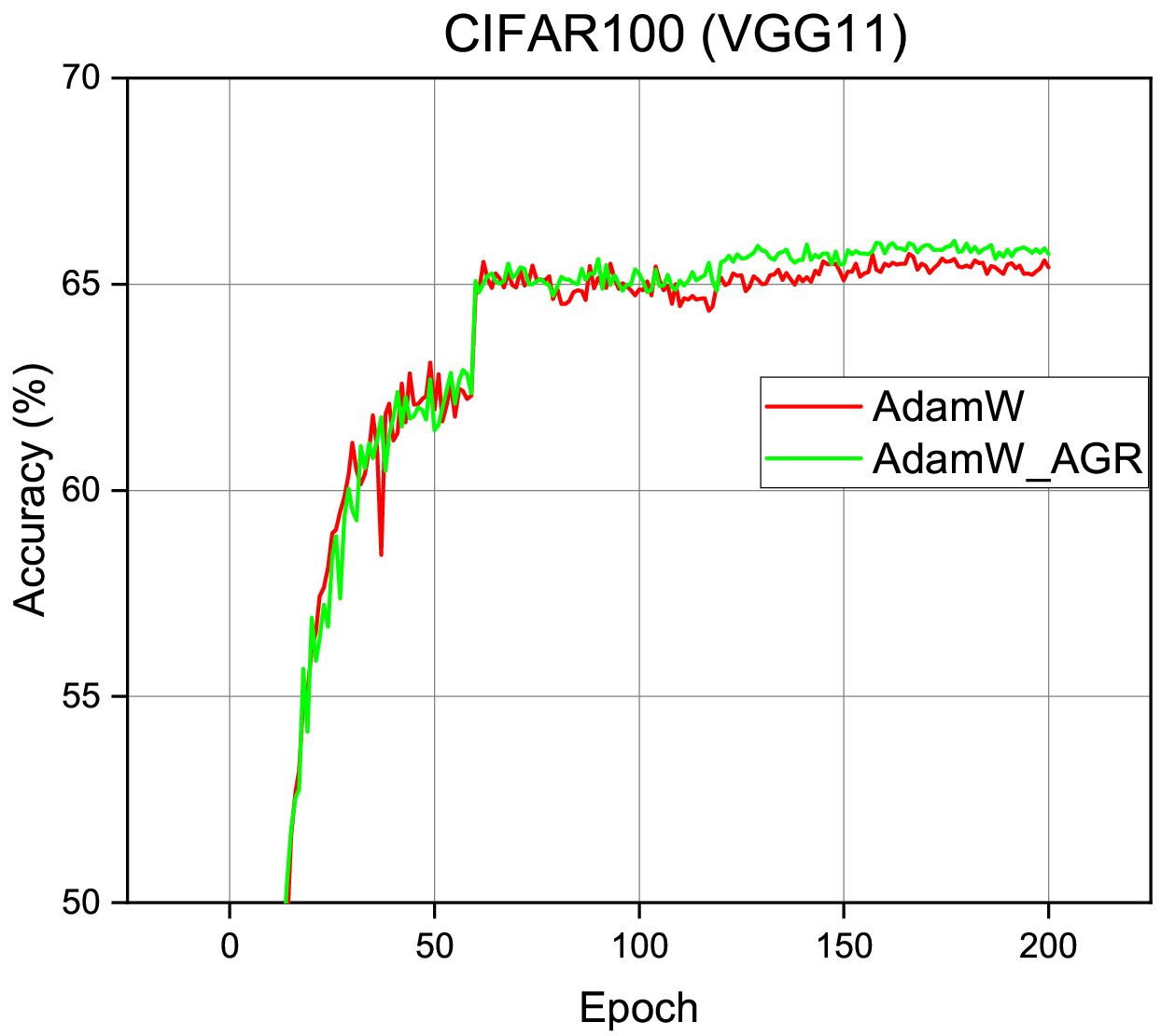}
    }	
    \quad
    \subfigure{
        \includegraphics[width=5.5cm,height=5.9cm]{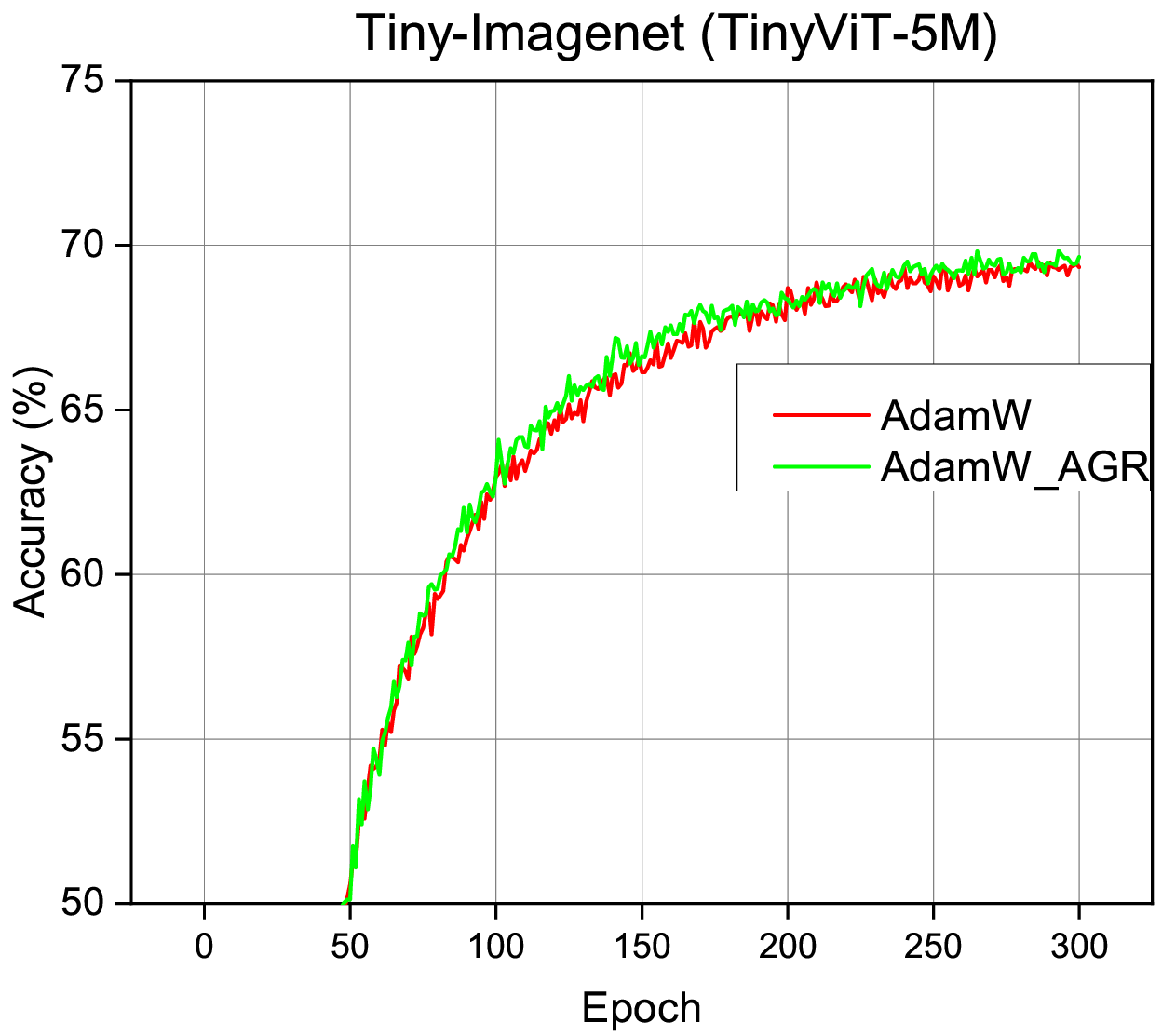}
    }
    \caption{Training loss and test accuracy in Tiny-Imagenet, AGR represents AGR is embedded into AdamW optimizer.}
    \vspace{-5mm}
\end{figure*}

\vspace{-1mm}
\subsection{5.4 Language Representations: ALBERT}
\noindent Here we applied the AGR method to train the light version of BERT: ALBERT. The language representation tasks consist of masked language model (MLM) and sentence order prediction (SOP). We evaluated the performance of the compared optimizers using the WikiText-2 dataset on a single GPU. Each of the experiments is conducted with 25 epochs. The batch size is 16. In this section, we also considered to evaluate the AGR method, by using multiple weight decays in AdamW. As shown in  Table 6, the AGR method can improve the SOP accuracy across different weight decays, and we also see the importance of weight decays in training the neural architectures.
\vspace{-2mm}
\subsection{5.5 Ablation Studies}
\noindent Through the extensive experiments, we found that the AGR method cannot improve the training performance during the entire process. According to our theoretical analysis, the gradient and learning rate will be reduced adaptively based on the current gradient magnitude. The learning rate will decrease more when the gradient magnitude is larger. However, the learning rate needs to be larger in the later training epochs. Hence, the performance will become better when suspending the AGR method after a period of training. Here, we conducted experiments on image classification tasks using the TinyViT-11M backbone, and image generation tasks using the DDPM, to evaluate the influence of the AGR over the epochs.
\noindent As shown in Table 7, when we ran the AGR method till the 250 epochs instead of the entire epochs, we can achieve higher accuracy. This validates our assumption that the learning rate in the later training epochs should be set to larger, relative to the epochs before. When we ran the AGR method with $1,200$ epochs instead of the entire training, we can obtain the best generalization performance.
\vspace{-3mm}
\subsection{6. Conclusions}
\vspace{-3mm}
This paper proposed a new optimization technique to improve the training efficiency and generalization performance for deep neural networks, in a simple-yet-effective way. In particular, theoretical results are provided to guarantee that the new algorithm can effectively smooth the loss landscape and adaptively adjust the learning rate. We conducted various deep neural network-related experiments to demonstrate the superior performance of the novel optimization algorithm, compared with the state-of-the-art optimization methods. In the near future, we will further explore the effectiveness of this novel optimization technique in accelerating the training efficiency and robustness for Transformer and Mamba, which may significantly benefit the research on large language models and video generations.

\bibliography{aaai25.bib}
\end{document}

\onecolumn 
\section{An Adaptive Gradient Regularization Method supplementary material}

\section{Dataset and experiment}
This work uses the CIFAR10 dataset for image generation and the CIFAR100 and Tiny-ImageNet datasets for image classification. In addition, we use the WikiText-2 dataset for language representation. We have shown the dataset partition in the paper. We want to stress that the Tiny-ImageNet dataset only has class labels in the training and validation sets. Hence, we use the validation set for evaluation. All images of the validation set are under the same directory instead of each class directory. Therefore, we process the validation set by classifying the images into different directories that conform to the same format as the training set.
For the hyperparameters in the training process, we use the default setting and optimizers of the original implementation. We use the official default setting for the optimizers, except that we consider the influence of weight decay in the language representation task. Because we mainly focus on the performance of the AGR method instead of higher accuracy by hyperparameters fine-tuning. We have zipped the code we implemented in the code.zip file. Our experiment can be reproducible using the default setting in the code repository.\\
Here, we represent the generation results at the 2000 epoch with and without the AGR method. The results show that the AGR method can improve image generation quality while accelerating the training process. For example, in the second row and the third column, the car generated with the AGR method is clearer; in the third row from the bottom and the last column, the face generated with the AGR method is clearer; in the fourth row and the third column from the bottom, the car generated with the AGR method is more intact; in the third row and the third column, the elk generated is much better with the AGR method.
\begin{figure*}[!ht] 
    \centering
    \subfigure[\label{fig:Adan}]{
        \includegraphics[width=8.3cm,height=8.3cm]{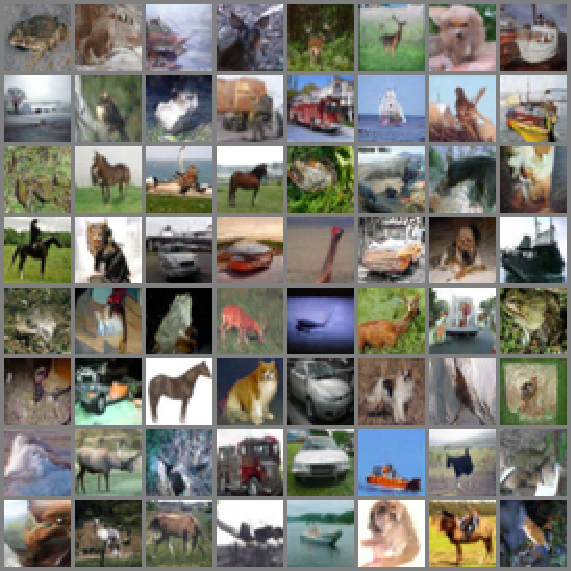}
    }	
        \quad
    \subfigure[\label{fig:Adan(AGR)}]{
        \includegraphics[width=8.3cm,height=8.3cm]{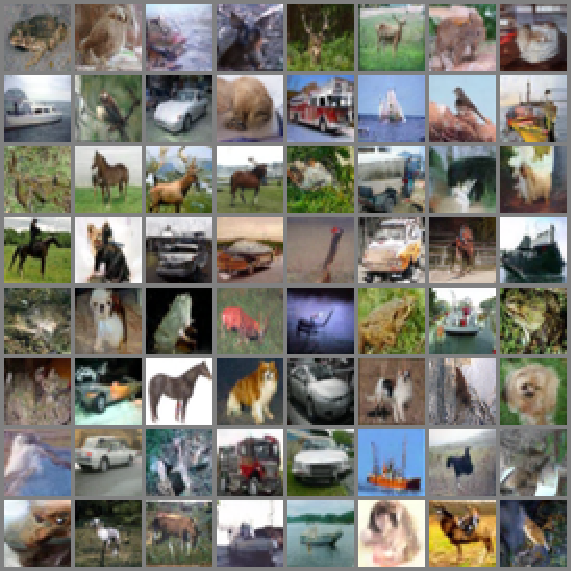}
    }
    \vspace{-1mm}
    \caption{Image generation at the 2000 epoch with (a) Adan and (b) Adan(AGR) }	
    \vspace{-6mm}
\end{figure*}
\section{Appendix}
\subsection{A1. Theorem 4.1 Proof}

\begin{proof}  
For weight $w_{i,j}$ in weight vectors and definition in eq(1) and eq(2), we have:
\begin{equation}
    \begin{split}
    \left|\left|\Psi(\nabla_{\textbf{w}_{i,j}}\mathcal{L})\right|\right|^{2}_{2} = &\left|\left|\nabla_{\textbf{w}_{i,j}}\mathcal{L}\right|\right|^{2}_{2}(1-\frac{\left| \nabla_{\textbf{w}_{i,j}} \mathcal{L} \right|}{\sum_{i,j}\left| \nabla_{\textbf{w}_{i,j}} \mathcal{L}\right|})^{2}\leq \left|\left|\nabla_{\textbf{w}_{i,j}}\mathcal{L}\right|\right|^{2}_{2}\\
    \end{split},
\end{equation}

\begin{equation}
    \begin{split}
        \nabla_{\textbf{w}_{i,j}}\Psi(\nabla_{\textbf{w}_{i,j}}\mathcal{L}) = &\nabla^2_{\textbf{w}_{i,j}}\mathcal{L}(1-\frac{\left| \nabla_{\textbf{w}_{i,j}} \mathcal{L} \right|}{\sum_{i,j}\left| \nabla_{\textbf{w}_{i,j}} \mathcal{L}\right|})^{2}+\nabla_{\textbf{w}_{i,j}}\mathcal{L} \nabla_{\textbf{w}_{i,j}}(1-\frac{\left| \nabla_{\textbf{w}_{i,j}} \mathcal{L} \right|}{\sum_{i,j}\left| \nabla_{\textbf{w}_{i,j}} \mathcal{L}\right|}).
        \\
    \end{split}
\end{equation}
Next, we need to discuss about two cases with $\nabla_{\textbf{w}_{i,j}}\mathcal{L}$\textless0 and $\nabla_{\textbf{w}_{i,j}}\mathcal{L}\geq0$.\\
when $\nabla_{\textbf{w}_{i,j}}\mathcal{L}\geq0$:

\begin{equation}
\begin{split}
\nabla_{\textbf{w}_{i,j}}(1-\frac{\left| \nabla_{\textbf{w}_{i,j}} \mathcal{L} \right|}{\sum_{i,j}\left| \nabla_{\textbf{w}_{i,j}} \mathcal{L}\right|})
     &=  -\nabla_{\textbf{w}_{i,j}}\frac{ \nabla_{\textbf{w}_{i,j}} \mathcal{L} }{\sum_{i,j}\left| \nabla_{\textbf{w}_{i,j}} \mathcal{L}\right|} \\
     &=  -\frac{\nabla_{\textbf{w}_{i,j}}^{2} \mathcal{L}\sum_{i,j}\left| \nabla_{\textbf{w}_{i,j}} \mathcal{L}\right|-\nabla_{\textbf{w}_{i,j}} \mathcal{L}\nabla_{\textbf{w}_{i,j}}\sum_{i,j}\left| \nabla_{\textbf{w}_{i,j}} \mathcal{L}\right|}{(\sum_{i,j}\left|\nabla_{\textbf{w}_{i,j}} \mathcal{L}\right|)^{2}} \\
     &=  -\frac{\nabla_{\textbf{w}_{i,j}}^{2} \mathcal{L}\sum_{i,j}\left| \nabla_{\textbf{w}_{i,j}} \mathcal{L}\right|-\nabla_{\textbf{w}_{i,j}} \mathcal{L}\nabla_{\textbf{w}_{i,j}}^{2} \mathcal{L}}{(\sum_{i,j}\left| \nabla_{\textbf{w}_{i,j}} \mathcal{L}\right|)^{2}}, \\
\end{split}
\end{equation}
Combined with eq(6), we have:\\
\begin{equation}
\begin{split}
\nabla_{\textbf{w}_{i,j}}\Psi(\nabla_{\textbf{w}_{i,j}}\mathcal{L})  &=\nabla^2_{\textbf{w}_{i,j}}\mathcal{L}(1-\frac{\left| \nabla_{\textbf{w}_{i,j}} \mathcal{L} \right|}{\sum_{i,j}\left| \nabla_{\textbf{w}_{i,j}} \mathcal{L}\right|})^{2}
     +  -\frac{\nabla_{\textbf{w}_{i,j}}^{2} \mathcal{L}\sum_{i,j}\left| \nabla_{\textbf{w}_{i,j}} \mathcal{L}\right|+\nabla_{\textbf{w}_{i,j}} \mathcal{L}\nabla_{\textbf{w}_{i,j}}^{2} \mathcal{L}}{(\sum_{i,j}\left| \nabla_{\textbf{w}_{i,j}} \mathcal{L}\right|)^{2}} \\
     &= \nabla^2_{\textbf{w}_{i,j}}\mathcal{L}\frac{(\sum_{i,j}\left|\nabla_{\textbf{w}_{i,j}} \mathcal{L}\right|)^{2}-2\nabla_{\textbf{w}_{i,j}\mathcal{L}}\sum_{i,j}\left| \nabla_{\textbf{w}_{i,j}} \mathcal{L}\right|+(\nabla{\textbf{w}_{i,j}}\mathcal{L})^{2}}{(\sum_{i,j}\left| \nabla_{\textbf{w}_{i,j}} \mathcal{L}\right|)^{2}}\\
     &= \nabla^2_{\textbf{w}_{i,j}}\mathcal{L}\frac{(\sum_{i,j}\left|\nabla_{\textbf{w}_{i,j}} \mathcal{L}\right|-\nabla_{\textbf{W}_{i,j}}\mathcal{L})^{2}}{(\sum_{i,j}\left|\nabla_{\textbf{w}_{i,j}} \mathcal{L}\right|)^{2}} \\
     &\leq \nabla^2_{\textbf{w}_{i,j}}\mathcal{L}.
\end{split}
\end{equation}
when $\nabla_{\textbf{w}_{i,j}}\mathcal{L}$\textless0:\\
\begin{equation}
\begin{split}
\nabla_{\textbf{w}_{i,j}}(1-\frac{\left| \nabla_{\textbf{w}_{i,j}} \mathcal{L} \right|}{\sum_{i,j}\left| \nabla_{\textbf{w}_{i,j}} \mathcal{L}\right|})
     &=  \nabla_{\textbf{w}_{i,j}}\frac{ \nabla_{\textbf{w}_{i,j}} \mathcal{L} }{\sum_{i,j}\left| \nabla_{\textbf{w}_{i,j}} \mathcal{L}\right|} \\
     &=  \frac{\nabla_{\textbf{w}_{i,j}}^{2} \mathcal{L}\sum_{i,j}\left| \nabla_{\textbf{w}_{i,j}} \mathcal{L}\right|-\nabla_{\textbf{w}_{i,j}} \mathcal{L}\nabla_{\textbf{w}_{i,j}}\sum_{i,j}\left| \nabla_{\textbf{w}_{i,j}} \mathcal{L}\right|}{(\sum_{i,j}\left| \nabla_{\textbf{w}_{i,j}} \mathcal{L}\right|)^{2}} \\
     &=  \frac{\nabla_{\textbf{w}_{i,j}}^{2} \mathcal{L}\sum_{i,j}\left| \nabla_{\textbf{w}_{i,j}} \mathcal{L}\right|+\nabla_{\textbf{w}_{i,j}} \mathcal{L}\nabla_{\textbf{w}_{i,j}}^{2} \mathcal{L}}{(\sum_{i,j}\left| \nabla_{\textbf{w}_{i,j}} \mathcal{L}\right|)^{2}}, \\
\end{split}
\end{equation}
Combined with eq(6), we have:\\
\begin{equation}
\begin{split}
\nabla_{\textbf{w}_{i,j}}\Psi(\nabla_{\textbf{w}_{i,j}}\mathcal{L})  &=\nabla^2_{\textbf{w}_{i,j}}\mathcal{L}(1+\frac{\nabla_{\textbf{w}_{i,j}} \mathcal{L}}{\sum_{i,j}\left| \nabla_{\textbf{w}_{i,j}} \mathcal{L}\right|})^{2}
     +  \frac{\nabla_{\textbf{w}_{i,j}}^{2} \mathcal{L}\sum_{i,j}\left| \nabla_{\textbf{w}_{i,j}} \mathcal{L}\right|+\nabla_{\textbf{w}_{i,j}} \mathcal{L}\nabla_{\textbf{w}_{i,j}}^{2} \mathcal{L}}{(\sum_{i,j}\left| \nabla_{\textbf{w}_{i,j}} \mathcal{L}\right|)^{2}} \\
     &= \nabla^2_{\textbf{w}_{i,j}}\mathcal{L}\frac{(\sum_{i,j}\left|\nabla_{\textbf{w}_{i,j}} \mathcal{L}\right|)^{2}+2\nabla_{\textbf{w}_{i,j}}\mathcal{L}\sum_{i,j}\left| \nabla_{\textbf{w}_{i,j}} \mathcal{L}\right|+(\nabla{\textbf{w}_{i,j}}\mathcal{L})^{2}}{(\sum_{i,j}\left| \nabla_{\textbf{w}_{i,j}} \mathcal{L}\right|)^{2}}\\
     &= \nabla^2_{\textbf{w}_{i,j}}\mathcal{L}\frac{(\sum_{i,j}\left|\nabla_{\textbf{w}_{i,j}} \mathcal{L}\right|+\nabla_{\textbf{W}_{i,j}}\mathcal{L})^{2}}{(\sum_{i,j}\left|\nabla_{\textbf{w}_{i,j}} \mathcal{L}\right|)^{2}} \\
     &\leq \nabla^2_{\textbf{w}_{i,j}}\mathcal{L}.
\end{split}
\end{equation}
In the optimization process, we need to find the optimal local minimum as the global minimum. Hence, the landscape of loss containing these local minima is convex. So we have:
\begin{equation}
\begin{split}
\left|\left|\nabla_{\textbf{w}_{i,j}}\Psi(\nabla_{\textbf{w}_{i,j}}\mathcal{L}) \right|\right|^{2}_{2}\leq \left|\left|\nabla^2_{\textbf{w}_{i,j}}\mathcal{L}\right|\right|^{2}_{2}.
\end{split}
\end{equation}
\end{proof}
\subsection{A2. Theorem 4.2 Proof}
\begin{proof}
In this proof, we take optimizer AdamW with AGR to illustrate theorem 4.2 is certified.
From the definition of momentum $m_{t}$ in AdamW at t-st update for a specific weight, we can derive recursively:
\begin{equation}
    m_{t} = \sum^{t-i}_{i=0}\beta^{i}_{1}(1-\beta_{1})(\Psi(\nabla_{\textbf{w}_{t}}\mathcal{L}))_{t-i}.
\end{equation}
From the simplicity we don't apply AGR to the second-order momentum, we have the following weight update process:
\begin{equation}
\begin{split}
    w_{t+1} &= w_{t}-\eta_{t} m_{t}\\
            &= w_{t}-\eta_{t} \sum^{t-i}_{i=0}\beta^{i}_{1}(1-\beta_{1})(\Psi(\nabla_{\textbf{w}_{t}}\mathcal{L}))_{t-i}\\
            &= w_{t}-\eta_{t} \sum^{t-i}_{i=0}\beta^{i}_{1}(1-\beta_{1})((1-\alpha_{t-i})(\nabla_{\textbf{w}_{t}}\mathcal{L}))_{t-i}.\\
\end{split}
\end{equation}
Assume for each time $\alpha_{t-i}=\hat{\alpha}$ is the same for clearer certification, we have:
\begin{equation}
\begin{split}
    w_{t+1} &= w_{t}-\eta_{t}\sqrt{(1-\hat{\alpha}^{t})} \sum^{t-i}_{i=0}\beta^{i}_{1}(1-\beta_{1})\sqrt{(1-\hat{\alpha}^{t})}(\nabla_{\textbf{w}_{t}}\mathcal{L})_{t-i}.\\
\end{split} 
\end{equation}
Through this perspective, we can see the same tendency in learning rate and gradient adaption with AGR.
\end{proof}